\documentclass[runningheads]{llncs}

\usepackage[final,year=2024,ID=7869]{eccv}

\usepackage{eccvabbrv}

\usepackage{graphicx}
\usepackage{booktabs}
\usepackage{multirow}
\usepackage{array}
\usepackage[accsupp]{axessibility}  

\usepackage{hyperref}

\usepackage{orcidlink}

\begin{document}

\title{YotoR-You Only Transform One Representation} 


\author{José Ignacio Díaz Villa\inst{1}\orcidlink{0009-0005-4531-2336} \and Patricio Loncomilla\inst{2}\orcidlink{1111-2222-3333-4444} \and
Javier Ruiz-del-Solar\inst{1,2}\orcidlink{2222--3333-4444-5555}}

\authorrunning{J. Díaz et al.}

\institute{Universidad de Chile, Santiago, Chile \and
AMTC, Santiago, Chile\\
\email{contacto@amtc.cl}}

\maketitle

\begin{abstract}
This paper introduces YotoR (You Only Transform One Representation), a novel deep learning model for object detection that combines Swin Transformers and YoloR architectures. Transformers, a revolutionary technology in natural language processing, have also significantly impacted computer vision, offering the potential to enhance accuracy and computational efficiency. YotoR combines the robust Swin Transformer backbone with the YoloR neck and head. In our experiments, YotoR models TP5 and BP4 consistently outperform YoloR P6 and Swin Transformers in various evaluations, delivering improved object detection performance and faster inference speeds than Swin Transformer models. These results highlight the potential for further model combinations and improvements in real-time object detection with Transformers. The paper concludes by emphasizing the broader implications of YotoR, including its potential to enhance transformer-based models for image-related tasks.
  \keywords{Object Detection \and Transformers \and Yolo}
\end{abstract}

\section{Introduction}
\label{sec:intro}


Convolutional neural networks have revolutionized computer vision applications in the last decade, enabling task-solving like object detection, image segmentation, and instance segmentation, among others. Despite the improvement of the convolutional network backbones in recent years, even surpassing human performance for several tasks, the use of Transformers \cite{transformers} in computer vision tasks remained elusive for several years.

The first application of transformers for computer vision tasks was proposed in 2020 \cite{visualtransformers}. However, because of the high resolution of images, the use of Transformers was limited to low-resolution applications like image classification. High-resolution tasks like object detection required the development of more specialized Transformers architectures like the Swin Transformer \cite{swintransformer}, which circumvent the computing limitations of Transformers by dynamically changing the attention window and allowing them to be used as a general purpose backbone for multiple vision tasks. Also, object detector heads based on transformers like DETR \cite{detr} have become state-of-the-art in tasks previously dominated by convolutional neural networks.

On the other hand, real-time object detectors, exemplified by the Yolo / YoloR family \cite{yolo} \cite{yolor}, remain indispensable for tasks dependent on high frame rates, such as autonomous driving, or on platforms constrained by limited hardware resources. Despite recent advancements in Transformers for computer vision, real-time object detection predominantly relies on convolutional neural networks. Their established reliability and computational efficiency in feature extraction have been a challenge for Transformers to overcome. Then, combining transformers with Yolo-like object detectors could deliver novel architectures able to achieve both high frame rates and high detection accuracies.

This paper introduces YotoR (You Only Transform One Representation), a novel deep learning model for object detection that combines Swin Transformers and YoloR architectures. YotoR combines the robust Swin Transformer backbone with the YoloR neck and head. In our experiments, YotoR models TP5 and BP4 consistently outperform YoloR P6 and Swin Transformers in various evaluations, delivering improved object detection performance and faster inference speeds than Swin Transformer models. These results highlight the potential for further model combinations and improvements in real-time object detection with Transformers.

The contributions of this paper are:

1) A new family of object detection architectures named YotoR, which is composed of backbones based on Swin Transformers and heads based on YoloR.

2) An exhaustive evaluation of different YotoR variants, which shows that YotoR models TP5 and BP4 consistently outperform YoloR P6 and Swin Transformers in various evaluations that consider both object detection performance and inference speed.

The code will be released upon acceptance of the paper.

\section{Related work}

\subsection{Real-time CNN-based object detection}

CNN-based object detectors, which started with Faster R-CNN \cite{fastrcnn}, have become a widely used approach to solving object detection tasks. In some applications, in which real-time processing is needed or there are hardware limitations, lightweight object detectors are needed.

The most widely used real-time object detectors are based on FCOS \cite{fcos} or the YOLO family of detectors \cite{yolo} \cite{yolov7}. Progress in this area has been made by improving the backbones, the necks, the detection heads, the loss functions, and the training procedures used. The backbone is the section of the network whose task is to obtain embeddings that encode the input images. In the case of the backbones, the improvements have been led by the use of building blocks inspired by RepVGG \cite{repvgg}, CSP \cite{cspnet}, ELAN \cite{elan}, and VoVNet \cite{vovnet}. The task of the neck is to further process the embeddings to make them useful for generating the detections. In the case of the necks, the most successful approaches are inspired by RepVGG \cite{repvgg} and PAN \cite{pan}. Finally, the detection heads generate the output detections of the network. In the case of the YOLO family of detectors, there are several heads, each one dedicated to detecting objects of different sizes. Then, a good design of the backbones, necks, and detection heads is critical for obtaining architectures capable of performing real-time detection without sacrificing performance.

\subsection{Multi-task-oriented real-time CNN architectures}

The use of multi-task architectures is promising, as they can integrate several information modalities for improving performance on all of the tasks. However, designing architectures able to perform multiple tasks in real time is challenging because using ensembles of per-task networks negatively affects the runtime of the system.

An architecture named YoloR \cite{yolor} was proposed to solve real-time multitask problems. The system is based on using a single unified architecture for solving several tasks. YoloR is based on modeling two sources of knowledge: explicit knowledge and implicit knowledge. Traditionally, explicit knowledge is strongly tied to the inputs of the network, and it is represented by the first layers of the network, while the dependence of the implicit knowledge on the inputs is lower, and it is encoded by the deeper layers. For achieving multi-task capabilities with a unified architecture, it is proposed that instead the explicit knowledge is modeled by weights shared by all of the input modalities, while the implicit knowledge is modeled as input-independent parameters in deep layers of the network. The weights related to the implicit knowledge can modify the internal network embeddings by applying operations like concatenation, addition, and multiplication. By using this approach, a single network architecture can benefit from being trained for multiple tasks, as the network weights related to the explicit knowledge can condense the information from all of the input modalities at the same time.

\begin{figure}[tb]
  \centering
  \begin{subfigure}{0.45\linewidth}
    \includegraphics[width=\columnwidth]{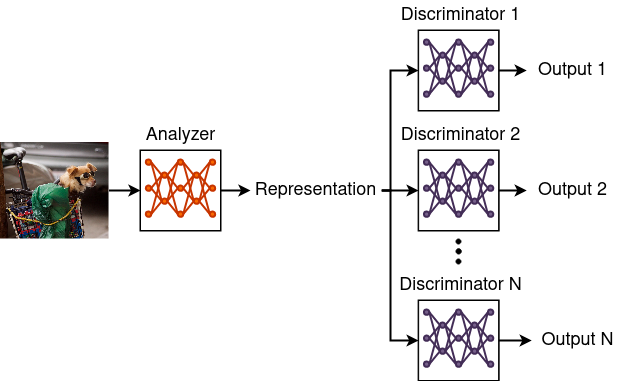}%
    \caption{Traditional multi-task architecture with a single analyzer.}
    \label{fig:short-a}
  \end{subfigure}
  \hfill
  \begin{subfigure}{0.45\linewidth}
    \includegraphics[width=\columnwidth]{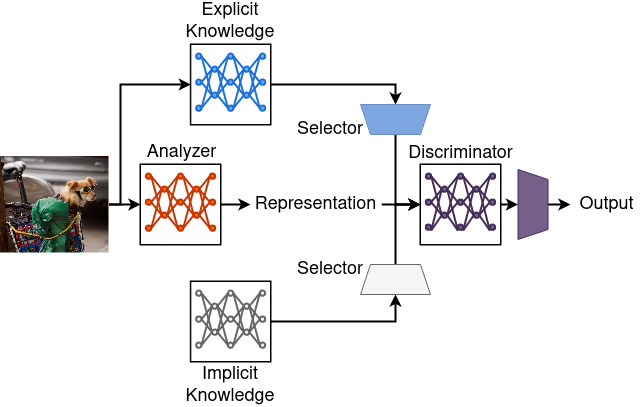}%
    \caption{Unified multi-task network with explicit and implicit knowledge.}
    \label{fig:short-b}
  \end{subfigure}
  \caption{Diagrams of the different approaches to a multi-task network as proposed by YoloR.}
  \label{fig:short}
\end{figure}

\subsection{Transformer-based object detection}

While transformers have been used successfully in natural language processing \cite{transformers}, their use in computer vision has been delayed because the amount of computing processing needed squares quadratically with the number of tokens (patches) in the image. The first work which used plain ViT (visual transformers) for classification \cite{visualtransformers} used patches of size 16x16 as tokens, and the transformer architecture consisted of a sequence of encoders. Despite the ViTs being able to process all of the tokens at the same time, they have no inductive biases like translation invariance, which are intrinsic to the CNNs architectures. Then, the largest ViT model proposed in that work needed over 300M images to achieve a better performance than state-of-the-art CNN models like ResNets trained with BiT \cite{bit}. Also, the problems of the quadratic increase in computational computing power by the number of tokens, the large size of the patches, and the lack of multiscale processing cause the plain ViT architecture to be unable to be used for tasks like object detection or image segmentation.

An efficient visual-transformer-based backbone able to replace CNN-based ones, named Swin Transformer, was proposed in \cite{swintransformer}. Swin transformers work by using small patches with size $4 \times 4$ as tokens and dividing the image into windows. Then, an encoder is applied to all of the tokens contained on each of the windows independently. This strategy enables the computing power requirement to remain low. As the windows are processed independently, their information must be combined in subsequent layers of the network. Then, after the application of encoders on the windows of a layer, the windows of the following layer are shifted. In other words, each window in a layer can integrate information from four windows in the previous layer. Also, after the application of a nonshifted and a shifted window layer, the tokens are merged, which increases the embedding dimension while decreasing the number of tokens. Also, the patch merging strategy enables the architecture to generate feature pyramids which are useful for tasks involving multiscale processing. Then, Swin transformers can be used instead of CNN-based backbones for tasks like object detection, object instance segmentation, and semantic segmentation.

In recent years, there has been a surge in proposals suggesting innovative combinations of architectures featuring Transformers for object detection. Notably, the authors of \cite{yoloxswin} present a distinctive fusion involving the anchor-free implementation of YOLO, namely YOLOX \cite{ge2021yolox}, and the Swin Transformer. An increase of 6.1\% in the mAP performance in their private dataset was observed for the detection of obstacles for autonomous driving, setting a good precedent for the combination of the Swin Transformer backbone with the Yolo family of detectors. 
Our proposal differs from the YOLOX-S models by keeping the YoloV3 anchors on the detection heads and implementing the improvements added by YoloV4 and YoloR like the new BoG and BoS techniques and the implicit knowledge modeling. Also, the proposed method was trained on the MSCOCO dataset to directly compare with the rest of the state-of-the-art. 

 Another strong contender has been the DETR \cite{detr} family of transformer-based object detection architectures, with some variants like the Co-DETR \cite{Zong2022DETRsWC} achieving state-of-the-art performance. The DETR works by using an encoder-decoder Transformer over features extracted by a backbone. This allows the model to achieve better accuracy and the ability to handle complex relations between the objects while keeping a simple structure for the architecture. 


\section{YotoR}

In this work, a family of network architectures is introduced that merges the Swin Transformer backbone with the YoloR head. Inspired by Yolo's nomenclature, these architectures are named YotoR, short for "You Only Transform One Representation". This reflects the use of a single unified representation generated by Transformer blocks, versatile and suitable for multiple tasks.
The idea behind this proposal is to use the powerful Swin Transformers feature extraction to improve detection accuracy, while also having the ability to solve multiple tasks with fast inference times by using the YoloR heads.

YotoR models following the YoloR architecture are named using the following convention: 

\begin{center}
    \textbf{YotoR \{Backbone Swin\}\{Head YoloR\}\{\# Blocks\} \\}
\end{center}

For a clearer understanding, consider the following examples:

\begin{itemize}
    \item YotoR TP5 has a Swin-T backbone, YoloR P6 head and neck, and a 5-block backbone.
    \item YotoR BP4 has a Swin-B backbone, YoloR P6 head and neck, and a 4-block backbone.
    \item YotoR LD4 has a Swin-L backbone, YoloR D6 head and neck, and a 4-block backbone.
\end{itemize}

Similar to the relationship between YoloR and its base model P6, YotoR TP4 is the starting point for YotoR models, representing the most basic combination of 
components.

Using the unaltered SwinT backbone also comes with a significant advantage, allowing for the application of transfer learning techniques. This is because, by not altering the Swin Transformer's structure, the publicly provided weights from its creators can be used. This simplifies the transfer of pre-trained Swin Transformer weights onto other datasets, speeding up the training process and improving performance.

\subsection{Architecture}

    \subsubsection{\textbf{Backbone:\\}}
    
\begin{figure}[ht]
    \hspace*{0.5cm}
    \centerline{\includegraphics[width=1.0\linewidth]{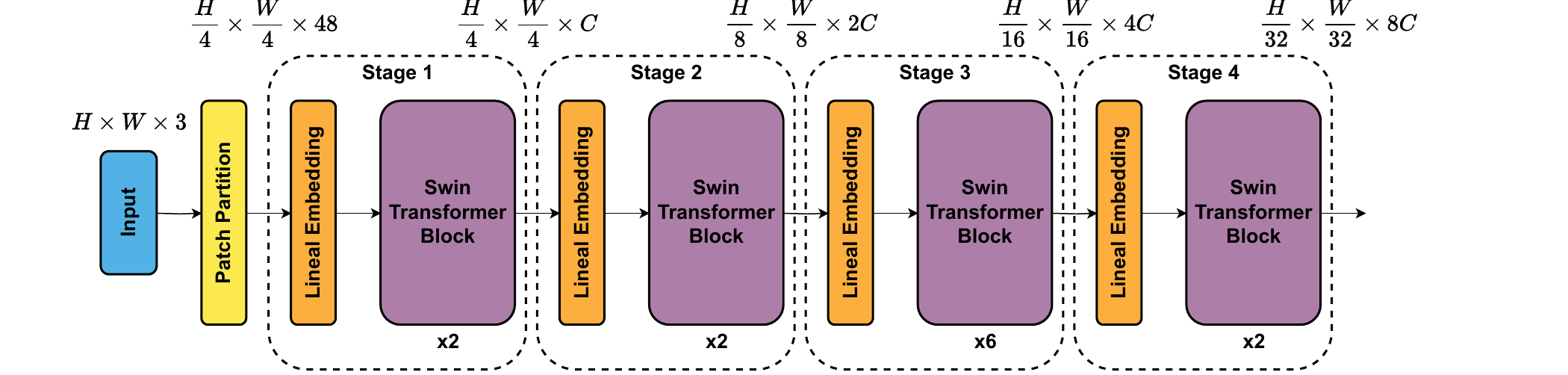}}
    \caption{Diagram of the Swin Transformer T backbone.}
    \label{img:SwinTModel}
    \end{figure}

    In \Cref{img:SwinTModel}, the Swin Transformer's simplified architecture used for the backbone is presented, particularly the T (tiny) version known as Swin T. The main part of the architecture has four stages; it starts by dividing the image into small $4 \times 4 \times 3$ patches, which are then transformed into tokens. These tokens go through a linear embedding layer to convert them into tokens with a certain size, referred to as $C$, which are then fed into the first Swin Transformer Block array that constitutes the first stage. The other three are composed of a patch merging block that merges patches by $2 \times 2$ groups, reducing the feature maps width and height resolution by half, which is then passed on to a corresponding array of Swin Transformer Blocks.

    



    
    Apart from Swin T, there are three more architectures: Swin S, Swin B, and Swin L. They are quite similar but differ in the number of Swin Transformer Blocks and channels. The details for each architecture can be found in \Cref{tab:SwinArchi}.
    
    For the object detection task in the Swin Transformer article \cite{swintransformer}, Mask R-CNN \cite{maskrcnn} and Cascade R-CNN \cite{cascadercnn} were the main heads used. In models B and L, they also used a framework named HTC++, which includes HTC \cite{htc}, instaboost \cite{instaboost}, and more advanced training methods with higher resolutions. While HTC++ gives better results, it is worth noting that it has not been made publicly available. Therefore, most of our focus will be on models without HTC++, unless we specify otherwise.

    \begin{table}[h!]
    \caption{Specification of the Swin Transformer architecture.}
    \label{tab:SwinArchi}
    \centering
    \small
        \begin{tabular}{c|c|cl|cl|cl|cl|}
                                 & \begin{tabular}[c]{@{}c@{}}output \\ size\end{tabular}            & \multicolumn{2}{c|}{\textbf{Swin-T}}                                                                   & \multicolumn{2}{c|}{\textbf{Swin-S}}                                                                    & \multicolumn{2}{c|}{\textbf{Swin-B}}                                                                     & \multicolumn{2}{c|}{\textbf{Swin-L}}                                                                     \\ \hline
        stage 1 & \multirow{2}{*}{\begin{tabular}[c]{@{}c@{}}4x\\ (56x56)\end{tabular}}  & \multicolumn{2}{c|}{\begin{tabular}[c]{@{}c@{}}concat 4x4, \\ 96-d, LN\end{tabular}}                   & \multicolumn{2}{c|}{\begin{tabular}[c]{@{}c@{}}concat 4x4, \\ 96-d, LN\end{tabular}}                    & \multicolumn{2}{c|}{\begin{tabular}[c]{@{}c@{}}concat 4x4, \\ 128-d, LN\end{tabular}}                    & \multicolumn{2}{c|}{\begin{tabular}[c]{@{}c@{}}concat 4x4, \\ 192-d, LN\end{tabular}}                    \\ \cline{3-10} 
                                 &                                                                        & \multicolumn{1}{c|}{\begin{tabular}[c]{@{}c@{}}{[}ws 7x7, \\ dim 96, \\ head 3{]}\end{tabular}}   & x2 & \multicolumn{1}{c|}{\begin{tabular}[c]{@{}c@{}}{[}ws 7x7, \\ dim 96, \\ head 3{]}\end{tabular}}   & x2  & \multicolumn{1}{c|}{\begin{tabular}[c]{@{}c@{}}{[}ws 7x7, \\ dim 128, \\ head 4{]}\end{tabular}}   & x2  & \multicolumn{1}{c|}{\begin{tabular}[c]{@{}c@{}}{[}ws 7x7, \\ dim 192, \\ head 6{]}\end{tabular}}   & x2  \\ \hline
        stage 2 & \multirow{2}{*}{\begin{tabular}[c]{@{}c@{}}8x\\ (28x28)\end{tabular}}  & \multicolumn{2}{c|}{\begin{tabular}[c]{@{}c@{}}concat 2x2, \\ 192-d, LN\end{tabular}}                  & \multicolumn{2}{c|}{\begin{tabular}[c]{@{}c@{}}concat 2x2, \\ 192-d, LN\end{tabular}}                   & \multicolumn{2}{c|}{\begin{tabular}[c]{@{}c@{}}concat 2x2, \\ 256-d, LN\end{tabular}}                    & \multicolumn{2}{c|}{\begin{tabular}[c]{@{}c@{}}concat 2x2, \\ 384-d, LN\end{tabular}}                    \\ \cline{3-10} 
                                 &                                                                        & \multicolumn{1}{c|}{\begin{tabular}[c]{@{}c@{}}{[}ws 7x7, \\ dim 192, \\ head 6{]}\end{tabular}}  & x2 & \multicolumn{1}{c|}{\begin{tabular}[c]{@{}c@{}}{[}ws 7x7, \\ dim 192, \\ head 6{]}\end{tabular}}  & x2  & \multicolumn{1}{c|}{\begin{tabular}[c]{@{}c@{}}{[}ws 7x7, \\ dim 256, \\ head 8{]}\end{tabular}}   & x2  & \multicolumn{1}{c|}{\begin{tabular}[c]{@{}c@{}}{[}ws 7x7, \\ dim 384, \\ head 12{]}\end{tabular}}  & x2  \\ \hline
        stage 3 & \multirow{2}{*}{\begin{tabular}[c]{@{}c@{}}16x\\ (14x14)\end{tabular}} & \multicolumn{2}{c|}{\begin{tabular}[c]{@{}c@{}}concat 2x2, \\ 384-d, LN\end{tabular}}                  & \multicolumn{2}{c|}{\begin{tabular}[c]{@{}c@{}}concat 2x2, \\ 384-d, LN\end{tabular}}                   & \multicolumn{2}{c|}{\begin{tabular}[c]{@{}c@{}}concat 2x2, \\ 512-d, LN\end{tabular}}                    & \multicolumn{2}{c|}{\begin{tabular}[c]{@{}c@{}}concat 2x2, \\ 768-d, LN\end{tabular}}                    \\ \cline{3-10} 
                                 &                                                                        & \multicolumn{1}{c|}{\begin{tabular}[c]{@{}c@{}}{[}ws 7x7, \\ dim 384, \\ head 12{]}\end{tabular}} & x6 & \multicolumn{1}{c|}{\begin{tabular}[c]{@{}c@{}}{[}ws 7x7, \\ dim 384, \\ head 12{]}\end{tabular}} & x18 & \multicolumn{1}{c|}{\begin{tabular}[c]{@{}c@{}}{[}ws 7x7, \\ dim 512, \\ head 16{]}\end{tabular}}  & x18 & \multicolumn{1}{c|}{\begin{tabular}[c]{@{}c@{}}{[}ws 7x7, \\ dim 768, \\ head 24{]}\end{tabular}}  & x18 \\ \hline
        stage 4 & \multirow{2}{*}{\begin{tabular}[c]{@{}c@{}}32x\\ (7x7)\end{tabular}}   & \multicolumn{2}{c|}{\begin{tabular}[c]{@{}c@{}}concat 2x2, \\ 768-d, LN\end{tabular}}                  & \multicolumn{2}{c|}{\begin{tabular}[c]{@{}c@{}}concat 2x2, \\ 768-d, LN\end{tabular}}                   & \multicolumn{2}{c|}{\begin{tabular}[c]{@{}c@{}}concat 2x2, \\ 1024-d, LN\end{tabular}}                   & \multicolumn{2}{c|}{\begin{tabular}[c]{@{}c@{}}concat 2x2, \\ 1536-d, LN\end{tabular}}                   \\ \cline{3-10} 
                                 &                                                                        & \multicolumn{1}{c|}{\begin{tabular}[c]{@{}c@{}}{[}ws 7x7, \\ dim 768, \\ head 24{]}\end{tabular}} & x2 & \multicolumn{1}{c|}{\begin{tabular}[c]{@{}c@{}}{[}ws 7x7, \\ dim 768, \\ head 24{]}\end{tabular}} & x2  & \multicolumn{1}{c|}{\begin{tabular}[c]{@{}c@{}}{[}ws 7x7, \\ dim 1024, \\ head 32{]}\end{tabular}} & x2  & \multicolumn{1}{c|}{\begin{tabular}[c]{@{}c@{}}{[}ws 7x7, \\ dim 1536,\\  head 48{]}\end{tabular}} & x2  \\ 
    \end{tabular}
    \end{table}
    
    \subsubsection{\textbf{Head}}
    
    To build the YoloR models, the authors decided to base them on the architecture of Scaled YoloV4 \cite{yolov4}. In particular, they started with YoloV4-P6-light as their foundation and sequentially modified it to create different versions of YoloR: P6, W6, E6, and D6. The changes between each of these versions are as follows:

    \begin{itemize}
        \item YoloR-P6: Replaced the Mish activation functions of YoloV4-P6-light with SiLU.
    
        \item YoloR-W6: Increased the number of channels in the outputs of the backbone blocks.
    
        \item YoloR-E6: Multiplied the number of channels from W6 by 1.25 and replaced the downsampling convolutions with CSP convolutions \cite{cspnet}.
    
        \item YoloR-D6: Increased the depth of the backbone.
    
    \end{itemize}

    
\subsubsection{\textbf{YotoR models}}

Selecting the YotoR models for implementation involved considering two essential aspects. Firstly, the discrepancy between the feature pyramid dimensions generated by the Swin Transformer backbone and the dimensions required by the YoloR head was analyzed. A significant disparity between these dimensions could create bottlenecks in the network, limiting its performance. 
 Second, to adapt the connections, the Swin Transformer features had to be re-shaped back to images with the attention maps. This is then normalized and passed through a $1 \times 1$ convolution to adjust the number of channels. This is done to give the YoloR head the same feature size as the DarknetCSP backbone \cite{yolov4} and to soften the information bottlenecks between the connections. 

Based on these considerations, we selected the YotoR TP4, YotoR TP5, YotoR BP4, and YotoR BB4 models, which are described below:

\begin{itemize}
    \item YotoR TP4 

    The smallest model among the proposed combinations, YotoR TP4, uses a Swin Transformer Tiny backbone with the YoloR P6 head and neck. However, this combination poses a challenge because the connections between these parts have drastically different dimensions, potentially leading to information bottlenecks. \\

    \item YotoR TP5 

    To address the bottleneck issue in YotoR TP4, it was decided to retain the B6 block, which matches the last CSPDarknet block in the YoloR backbone. This choice allowed for more coherent connections between both parts, preventing information loss in the transition to the neck. The "5" in its name refers to the inclusion of the additional B6 CSPDarknet block in addition to the four Swin Transformer blocks. \\
    
    \item YotoR BP4 
    
    It follows a similar premise and structure as YotoR TP4, but it replaces the Swin Transformer Tiny (T) backbone with a Swin Transformer Basic (B) backbone. The decision to opt directly for the Basic backbone, rather than the Small (S) backbone, is based on the similarity in connection sizes with the YoloR P6 backbone to avoid connection discrepancies, as the case in YotoR TP4. \\

    \item YotoR BB4 

    As the final proposed model, YotoR BB4 takes a different approach to its design. Instead of adapting the connections between the YoloR P6 neck and the Swin B backbone, we decided to completely redesign the P6 head, following the dimensions of the Swin B embeddings. This results in an expansion of dimensions in the deeper parts of the network while reducing them in the shallower layers. In YotoR BB4, all model dimensions are directly in accordance, with no adapting convolutional layers, unlike the previous models. \\

    \end{itemize}

\begin{figure}[htbp]
\centerline{\includegraphics[width=0.8\linewidth]{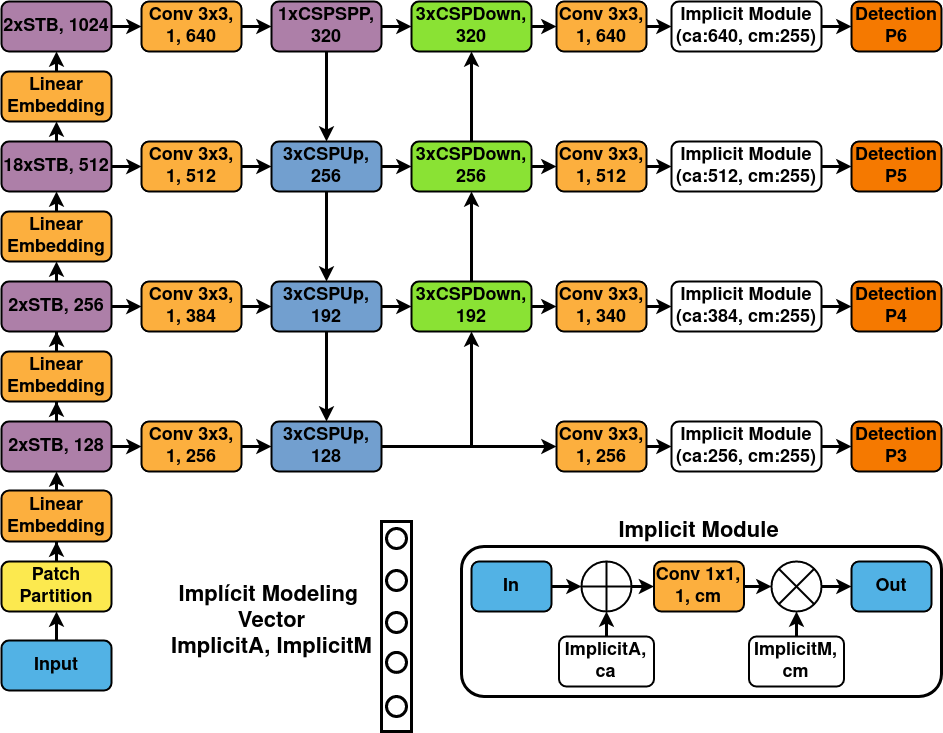}}
\caption{Architecture of YotoR BP4. }
\label{fig:yotorbp4}
\end{figure}

\Cref{fig:yotorbp4} shows the YotoR BP4 architecture. It introduces the STB (Swin Transformer Block), representing the Swin Transformer blocks used in the different YotoR architectures. Additionally, there is an inclusion of a Linear Embedding block between these components. This Linear Embedding block comes from the Swin Transformer implementation for object detection and is incorporated into the YotoR implementation without alterations.

These four models were chosen because they are composed of the base architectures of YoloR and Swin Transformer, allowing for a valid comparison to assess the proposed model's validity. While training and evaluating larger models like YotoR BW4 or YotoR BW5 were considered, resource limitations on a V100 GPU made this option unfeasible.

\section{Experimental Results}

The experiments to test and compare the models were implemented on the MSCOCO dataset. This dataset was chosen because of its common use for benchmarking object detectors.

\subsection{Experimental Setup}

Both training and evaluation were performed on a V100 GPU with 32 GB and 16 GB respectively. 
\footnote{All provided speed evaluations of the implemented models were performed on a similar V100 GPU with 16 GB and no TensorRT, and are therefore not directly comparable with those reported by other groups on COCO val2017
}
The training parameters used are the same as YoloR as shown in \Cref{tab:trainingparameters}. The YoloR training schedule was followed and 
trained for 300 epochs first, then fine-tuned using 125 epochs, and finally trained for an extra 25 epochs using a lower mosaic count.

\begin{table}[tb]
\caption{Training parameters.}
\label{tab:trainingparameters}
\centering
\scriptsize
\begin{tabular}{l|ccc}
Parameter                         & Training                & Tuning  & End                    \\ \hline
num epochs                        & 300 & 125 & 25 \\
initial learning rate             & 0.01                    & 0.01                    & 0.01                   \\
final learning rate               & 0.2                     & 0.2                     & 0.2                    \\
momentum                          & 0.937                   & 0.937                   & 0.937                  \\
optimizer weight decay            & 0.0005                  & 0.0005                  & 0.0005                 \\
warmup epochs                     & 3.0                     & 3.0                     & 3.0                    \\
warmup initial momentum           & 0.8                     & 0.8                     & 0.8                    \\
warmup initial bias lr            & 0.1                     & 0.1                     & 0.1                    \\
box loss gain                     & 0.05                    & 0.05                    & 0.05                   \\
cls loss gain                     & 0.5                     & 0.5                     & 0.5                    \\
cls BCELoss positive\_weight      & 1.0                     & 1.0                     & 1.0                    \\
obj loss gain                     & 1.0                     & 1.0                     & 1.0                    \\
obj BCELoss positive weight       & 1.0                     & 1.0                     & 1.0                    \\
IoU training threshold            & 0.2                     & 0.2                     & 0.2                    \\
anchor-multiple threshold         & 4.0                     & 4.0                     & 4.0                    \\
focal loss gamma                  & 0.0                     & 0.0                     & 0.0                    \\
image HSV-Hue augmentation        & 0.015                   & 0.015                   & 0.015                  \\
image HSV-Saturation augmentation & 0.7                     & 0.7                     & 0.7                    \\
image HSV-Value augmentation      & 0.4                     & 0.4                     & 0.4                    \\
image rotation                    & 0.0                     & 0.0                     & 0.0                    \\
image translation                 & 0.5                     & 0.5                     & 0.5                    \\
image scale                       & 0.5                     & 0.8                     & 0.8                    \\
image shear                       & 0.0                     & 0.0                     & 0.0                    \\
perspective                       & 0.0                     & 0.0                     & 0.0                    \\
image flip up-down                & 0.0                     & 0.0                     & 0.0                    \\
image flip left-right             & 0.5                     & 0.5                     & 0.5                    \\
image mosaic probability          & 1.0                     & 1.0                     & 1.0                    \\
image mosaic quantity             & 4.0                     & 9.0                     & 4.0                    \\
image mixup                       & 0.0                     & 0.2                     & 0.2                   
\end{tabular}
\end{table}


\subsection{Results of Inference Speed}

From the results shown in \Cref{tab:val2017results}, it is clear that none of the YotoR models can match the speed of YoloR. However, all the YotoR models have increased inference speeds compared to their respective Swin backbones. In particular, TP5 is notable for its inference frame rate that more than doubles compared to Swin-T, while BP4 and BB4 increase by just over 80\% to Swin B.


Due to significant variation in time measurements across models, we conducted all timing evaluations on a consistent platform: a 16GB V100 GPU, operating at a resolution of $1280 \times 1280$, without non-maximum suppression (NMS) or TensorRT acceleration. Each timing measurement represents the mean of three consecutive runs performed sequentially within a single session. Despite our best efforts, the recorded times were approximately half of those reported in the papers for YoloR and Swin Transformer models. This suggests that there is potential to double the speed results through further optimization. 

\begin{table}[] 
\caption{Results of the inference speed evaluation.}
\label{tab:resultsfps}
\centering
\small
\begin{tabular}{l|l|l}
model             & inference time (ms) & frames per second (fps) \\ \hline
Swin T & 135.1                          & 7.4                        \\
Swin B & 196.1 & 5.1                        \\
YoloR P6           & 34.7& 28.8\\
YotoR TP4          & 51.9& 19.3\\
YotoR TP5          & 48.6& 20.6\\
YotoR BP4          & 106.2& 9.4\\
YotoR BB4          & 103.9& 9.6\end{tabular}
\end{table}

\subsection{Results on COCO val2017}

The mAP (mean Average Precision) of the four YotoR models on val2017 is shown in  \Cref{tab:val2017results}. TP5 and BP4 models outperform all the baselines, even YoloR P6, which serves as a reference due to its high mAP performance. The only model that does not surpass it is BB4. However, considering that BB4 is built solely from Swin B without using the YoloR P6 head, this result was expected. However, BB4 manages to outperform Swin B in terms of performance, suggesting the possibility of exploring solutions in larger models such as Swin L, which surpasses the performance of YoloR D6.

\begin{figure}[h!]
\centering
  \includegraphics[width=\linewidth]{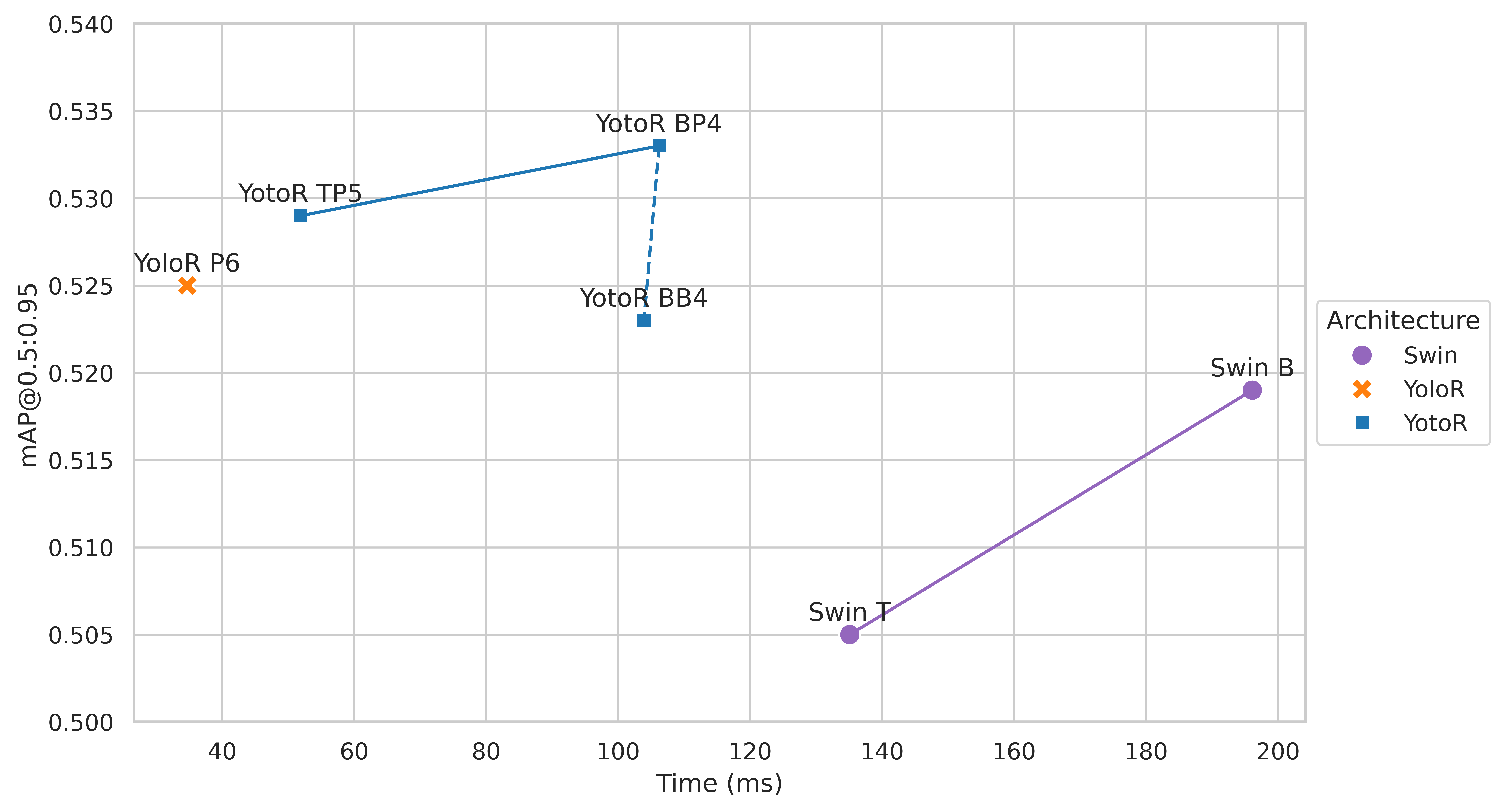}
\caption{Comparison between the time and mAP of each model in COCO val2017.}
\label{fig:mapval2017}
\end{figure}

The advantages of YotoR models become evident when comparing inference time in milliseconds with mAP, as shown in \Cref{fig:mapval2017}. It allows us to analyze the relationship between inference speed and model performance, which is essential because there is typically a trade-off between both. Speed tends to decrease when aiming for higher accuracy, and vice versa. In this figure, it is clear how all the YotoR models outperform their equivalent Swin detectors in both performance and speed, representing a significant improvement for all these models. Although YoloR P6 still has a considerable advantage in inference time, its performance can be surpassed by YotoR TP5 and YotoR BP4, despite using the same head architecture.

\begin{table*}[h!]
\caption{Evaluation results in val2017.}
\label{tab:resultsval2017}
\centering
\tiny
\begin{tabular}{llll|l|l|l|l|l|l|l}
                        &                                    &                                  &             & Swin T & Swin B         & YoloR P6       & YotoR TP4 & YotoR TP5 & YotoR BP4      & YotoR BB4 \\
\multicolumn{1}{l|}{AP} & \multicolumn{1}{l|}{IoU=.5:.95} & \multicolumn{1}{l|}{area=all} & maxDets=100 & 0.505  & 0.519          & 0.525          & 0.234     & 0.529     & \textbf{0.533} & 0.523     \\
\multicolumn{1}{l|}{AP} & \multicolumn{1}{l|}{IoU=.5}      & \multicolumn{1}{l|}{area=all} & maxDets=100 & 0.693  & 0.705          & 0.707          & 0.555     & 0.712     & \textbf{0.720} & 0.713     \\
\multicolumn{1}{l|}{AP} & \multicolumn{1}{l|}{IoU=.75}      & \multicolumn{1}{l|}{area=all} & maxDets=100 & 0.549  & 0.564          & 0.575          & 0.163     & 0.580     & \textbf{0.583} & 0.569     \\
\multicolumn{1}{l|}{AP} & \multicolumn{1}{l|}{IoU=.5:.95} & \multicolumn{1}{l|}{area=s} & maxDets=100 & -      & 0.354          & 0.371          & 0.138     & 0.384     & \textbf{0.388} & 0.369     \\
\multicolumn{1}{l|}{AP} & \multicolumn{1}{l|}{IoU=.5:.95} & \multicolumn{1}{l|}{area=m} & maxDets=100 & -      & 0.552          & 0.569          & 0.251     & 0.576     & \textbf{0.577} & 0.567     \\
\multicolumn{1}{l|}{AP} & \multicolumn{1}{l|}{IoU=.5:.95} & \multicolumn{1}{l|}{area=l} & maxDets=100 & -      & 0.673          & 0.661          & 0.352     & 0.661     & \textbf{0.675} & 0.673     \\
\multicolumn{1}{l|}{AR} & \multicolumn{1}{l|}{IoU=.5:.95} & \multicolumn{1}{l|}{area=all} & maxDets=  1 & -      & \textbf{0.638} & 0.392          & 0.196     & 0.391     & 0.390          & 0.388     \\
\multicolumn{1}{l|}{AR} & \multicolumn{1}{l|}{IoU=.5:.95} & \multicolumn{1}{l|}{area=all} & maxDets= 10 & -      & 0.638          & \textbf{0.652} & 0.433     & 0.643     & 0.639          & 0.631     \\
\multicolumn{1}{l|}{AR} & \multicolumn{1}{l|}{IoU=.5:.95} & \multicolumn{1}{l|}{area=all} & maxDets=100 & -      & 0.638          & \textbf{0.714} & 0.523     & 0.699     & 0.693          & 0.683     \\
\multicolumn{1}{l|}{AR} & \multicolumn{1}{l|}{IoU=.5:.95} & \multicolumn{1}{l|}{area=s} & maxDets=100 & -      & 0.478          & \textbf{0.578} & 0.373     & 0.567     & 0.553          & 0.534     \\
\multicolumn{1}{l|}{AR} & \multicolumn{1}{l|}{IoU=.5:.95} & \multicolumn{1}{l|}{area=m} & maxDets=100 & -      & 0.673          & \textbf{0.753} & 0.576     & 0.738     & 0.732          & 0.721     \\
\multicolumn{1}{l|}{AR} & \multicolumn{1}{l|}{IoU=.5:.95} & \multicolumn{1}{l|}{area=l} & maxDets=100 & -      & 0.782          & \textbf{0.840} & 0.648     & 0.820     & 0.818          & 0.818    
\end{tabular}
\end{table*}

\subsection{Results on COCO testdev}

Looking at \Cref{tab:testdevresults}, it can be observed that the YotoR models TP5 and BP4 outperform the YoloR P6 model in all mAP metrics. This confirms the significant improvement in the performance of these models. These results are publicly available thanks to the CodaLab team, making it easy to compare them with other participants in the COCO competition.

In addition to the traditional metrics, \Cref{tab:testdevresults} includes measurements of FLOPs (Floating-Point Operations) and the number of parameters for each model, which provide interesting insights. YotoR models have fewer FLOPs than YoloR P6, even though they have longer inference times. This could be explained by the nature of floating-point operations in YotoR, which, although less numerous, might be more computationally intensive or less optimized than those in YoloR. This suggests the potential for optimizing the inference time of YotoR models by improving their operations. On the other hand, the number of parameters in YotoR models falls in between Swin and YoloR models, as expected, since they combine elements from both.

These results are promising, especially when considering that the YotoR models were implemented using the most basic versions of YoloR and the Swin Transformer. This opens the door to exploring new combinations between models from the YOLO family and Swin Transformer, either to create a real-time object detection model with Transformers or to advance the state of the art in performance.

\begin{table*}[h] 
\caption{Results comparing the YotoR models and the \textit{baselines}.}
\label{tab:val2017results}
\centering
\tiny
\begin{tabular}{l|llll|llll|llll}
         & \multicolumn{4}{l|}{YotoR Tp5}                                                                             & \multicolumn{4}{l|}{YotoR Bp4}                                                                            & \multicolumn{4}{l}{YotoR BB4}                                                                             \\ \cline{2-13} 
         & mAP.5:.95 & \multicolumn{1}{l|}{}                               & FPS  &                                 & mAP.5:.95 & \multicolumn{1}{l|}{}                               & FPS  &                                & mAP.5:.95 & \multicolumn{1}{l|}{}                               & FPS  &                                \\
         & \multicolumn{2}{l|}{52.9\%}                                       & \multicolumn{2}{l|}{20.6}              & \multicolumn{2}{l|}{53.6\%}                                       & \multicolumn{2}{l|}{9.4}              & \multicolumn{2}{l|}{52.3\%}                                       & \multicolumn{2}{l}{9.6}               \\ \hline
Swin T   & 50.4\%      & \multicolumn{1}{l|}{{\color[HTML]{32CB00} +2.5\%}} & 7.4  & {\color[HTML]{32CB00} +178\%}    & 50.4\%      & \multicolumn{1}{l|}{{\color[HTML]{32CB00} +3.2\%}} & 7.4  & \multicolumn{1}{l|}{{\color[HTML]{32CB00} +27\%}}   & 50.4\%      & \multicolumn{1}{l|}{{\color[HTML]{32CB00} +1.9\%}} & 7.4  & {\color[HTML]{32CB00} +30\%}  \\
Swin B   & 51.9\%      & \multicolumn{1}{l|}{{\color[HTML]{32CB00} +1.0\%}} & 5.1  & {\color[HTML]{32CB00} +303\%}   & 51.9\%      & \multicolumn{1}{l|}{{\color[HTML]{32CB00} +1.7\%}} & 5.1  & \multicolumn{1}{l|}{{\color[HTML]{32CB00} +84\%}}                        & 51.9\%      & \multicolumn{1}{l|}{{\color[HTML]{32CB00} +0.4\%}} & 5.1  & {\color[HTML]{32CB00} +88\%}  \\
YoloR P6 & 52.5\%      & \multicolumn{1}{l|}{{\color[HTML]{32CB00} +0.4\%}} & 28.8 & {\color[HTML]{FE0000} - 28\%} & 52.5\%      & \multicolumn{1}{l|}{{\color[HTML]{32CB00} +1.1\%}} & 28.8 & {\color[HTML]{FE0000} - 70\%} & 52.5\%      & \multicolumn{1}{l|}{{\color[HTML]{FE0000} - 0.3\%}} & 28.8 & {\color[HTML]{FE0000} - 67\%}
\end{tabular}%
\end{table*}

\begin{table*}[h]
\caption{Comparison of YotoR in testdev}
\label{tab:testdevresults}
\centering
\scriptsize
\begin{tabular}{lllllllllll}
\hline
Model    & Size & FPS  & FLOPs & \# parameters & $AP^{test}$ & $AP^{test}_{50}$ & $AP^{test}_{75}$ & $AP^{test}_{S}$ & $AP^{test}_{M}$ & $AP^{test}_{L}$ \\ \hline
YoloR P6  & 1280 & 28.8 & 326G  & 37M           & 52.6\%                       & 70.6\%                               & 57.6\%                               & 34.7\%                              & 56.6\%                              & 64.2\%                              \\
Swin T    & 1280 & 7.4  & 745G & 86M          & -                            & -                                    & -                                    & -                                   & -                                   & -                                   \\
Swin B    & 1280 & 5.1  & 982G & 145M         & -                            & -                                    & -                                    & -                                   & -                                   & -                                   \\
YotoR TP5 & 1280 & 20.6 & 107G  & 56M           & 52.9\%                       & 70.9\%                               & 57.7\%                               & \textbf{35.3\%}                     & 56.5\%                              & 64.7\%                              \\
YotoR BP4 & 1280 & 9.4  & 296G  & 117M          & \textbf{53.4\%}              & \textbf{71.8\%}                      & \textbf{58.2\%}                      & 35.1\%                              & \textbf{57.1\%}                     & \textbf{66.5\%}                     \\
YotoR BB4 & 1280 & 9.6  & 281G  & 131M          & 52.4\%                       & 71.0\%                               & 57.1\%                               & 33.8\%                              & 56.1\%                              & 65.7                               
\end{tabular}%
\end{table*}

\begin{figure}[htbp]
\centerline{\includegraphics[width=\linewidth]{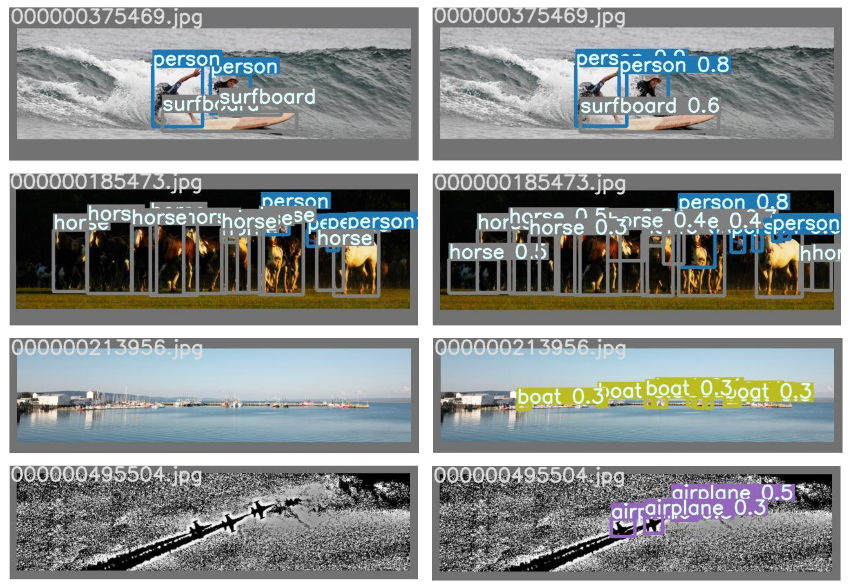}}
\caption{Left: Images from val2017 and testdev. Right: Predictions from YotoR BP4.}
\label{fig:predictions}
\end{figure}

In \Cref{fig:predictions}, we showcase instances of detections executed by the YotoR BP4 model on the val2017 and testdev datasets. These examples demonstrate the model's proficiency in effectively detecting objects of varying sizes, even when they overlap within a group.

\subsection{Comparison with the state of the art}

In \Cref{tab:sotacomparison}, we present a detailed comparison between the YotoR models and various state-of-the-art object detection models, which encompass different sizes and architectures. Our focus specifically centers on models of similar sizes and structures. This table provides insights into the performance of these models in both the COCO val2017 and testdev datasets.

While it is evident that the YotoR models exhibit lower performance compared to some of the latest and larger models, it is important to note that only the smaller versions of YotoR were implemented, featuring the smallest number of channels in the connections. This choice accounts for their comparatively lower performance compared to YoloR-W6, YOLOv7-W6, and their larger variants. Despite this, the YotoR models still outperform its individual components like YoloR-P6 and Swin Transformer B. It is worth mentioning that the performance surpasses that of the Swin Transformer without HTC++, albeit with certain details not publicly disclosed and some adaptations made to the YoloR head for processing images of varying resolutions.

In essence, the YotoR models appear to strike a commendable balance between performance and speed. It's noteworthy that this equilibrium could potentially be further optimized by incorporating larger models within the YotoR family, such as YotoR LE5 or YotoR LD5. The implementation of these larger variants holds promise for enhancing the competitiveness of the models, opening up intriguing avenues for future exploration.


\begin{table*}[h]
\caption{Comparison with the state of the art (batch=1, GPU=V100). * indicates our own confirmed measurements  using a 16GB V100 GPU .}
\tiny
\centering
\begin{tabular}{lllllllll}
\hline
model                                                     & \# parameters & FLOPs  & size            & $FPS^{V100}$ & $AP^{val}$ & $AP^{test}$ & $AP^{test}_{50}$ & $AP^{test}_{75}$ \\ \hline
YOTOR-TP4                                                 & 55M           & 111.5G & 1280            & 19.3*                          & 23.9\%     & 23.5\%      & 54.9\%           & 15.7\%           \\
YOTOR-TP5                                                 & 56M           & 107G   & 1280            & 20.6*                            & 52.9 \%    & 52.9 \%     & 70.9 \%          & 57.7 \%          \\
YOTOR-BP4                                                 & 117M          & 296G   & 1280            & 9.4*                            & 53.6 \%    & 53.4 \%     & 71.8 \%          & 58.2 \%          \\
YOTOR-BB4                                                 & 131M          & 281G   & 1280            & 9.6*                          & 52.3 \%    & 52.4 \%     & 71.0 \%          & 57.1 \%          \\ \hline
YOLOR-P6                                                  & 37M           & 326G   & 1280            & 76/28.8*                            & 52.5\%     & 52.6\%      & 70.6\%           & 57.6\%           \\
YOLOR-P6D                                                 & 37M           & 326G   & 1280            &                               & -          & 53.0\%      & 71.0\%           & 58.0\%           \\
YOLOR-W6                                                  & 80M           & 454G   & 1280            & 66                            & 54.0\%     & 54.1\%      & 72.0\%           & 59.2\%           \\
YOLOR-E6                                                  & 116M          & 684G   & 1280            & 45                            & 54.6\%     & 54.8\%      & 72.7\%           & 60.0\%           \\
YOLOR-D6                                                  & 152M          & 937G   & 1280            & 34                            & 55.4\%     & 55.4\%      & 73.3\%           & 60.6\%           \\ \hline
YOLOv7                                                    & 36.9M         & 104.7G & 640             & 161                              & 51.2\%     & 51.4\%      & 69.7\%           & 55.9\%           \\
YOLOv7-W6                                                 & 70.4M         & 360.0G & 1280            &  84                             & 54.6\%     & 54.9\%      & 72.6\%           & 60.1\%           \\
YOLOv7-E6                                                 & 97.2M         & 515.2G & 1280            &  56                             & 55.9\%     & 56.0\%      & 73.5\%           & 61.2\%           \\
YOLOv7-D6                                                 & 154.7M        & 806.8G & 1280            &  44                             & 56.3\%     & 56.6\%      & 74.0\%           & 61.8\%           \\ \hline
YOLOv4-CSP-P5 \cite{yolov4}              & 71M           & 328G   & 896             &  41                             & 51.7\%           & 51.8\%      & 70.3\%           & 56.6\%           \\
YOLOv4-CSP-P6 \cite{yolov4}              & 128M          & 718G   & 1280            &  30                             & 54.4\%     & 54.5\%      & 72.6\%           & 59.8\%           \\
YOLOv4-CSP-P7 \cite{yolov4}              & 287M          & 1639G  & 1536            &  16                             &  55.3\%          & 55.5\%      & 73.4\%           & 60.8\%           \\ \hline
YOLOv6-L6 \cite{li2023yolov6}            & 140M          & 773.4G & -               & 26                              & 57.2\%     & -           & -                & -                \\
PRB-FPN6-E-ELAN \cite{chen2021parallel}  & -             & -      & -               & 31                              & -          & 56.9\%      & 74.1\%           & 62.7\%           \\ \hline
Swin T-T (C-M-RCNN)                                       & 86M           & 745G   & -               & 15.3/7.4*                         & 50.5\%     & -           & -                & -                \\
Swin T-S (C-M-RCNN)                                       & 107M          & 838G   & -               & 12                            & 51.8\%     & -           & -                & -                \\
Swin T-B (C-M-RCNN)                                       & 145M          & 982G   & -               & 11.6/5.1*                         & 51.9\%     & -           & -                & -                \\
Swin T-B (HTC++)                                          & 160M          & 1043G  & -               & 11.6                          & 56.4\%     & -           & -                & -                \\
Swin T-L (HTC++)                                          & 284M          & 1470G  & -               & -                             & 57.1\%     & 57.7\%      & -                & -                \\ \hline
SwinV2-L(HTC++) \cite{swinv2}            & 197M          & -      & 1536            & -                             & 60.2\%     & 60.8\%      & -                & -                \\
SwinV2-G(HTC++) \cite{swinv2}            & 3.0B          & -      & 1536            & -                             & 62.5\%     & 63.1\%      & -                & -                \\ \hline
DETR \cite{detr}                         & 41M           & 86G    & \textless{}1333 &                               & 42.0\%     &             &                  &                  \\
DETR-DC5 \cite{detr}                     & 41M           & 187G   & \textless{}1333 &                               & 43.3\%     &             &                  &                  \\
DETR-R101 \cite{detr}                    & 60M           & 152G   & \textless{}1333 &                               & 43.5\%     &             &                  &                  \\
DETR-DC5-R101 \cite{detr}                & 60M           & 253G   & \textless{}1333 &                               & 44.9\%     &             &                  &                  \\ \hline
Co-DETR       \cite{Zong2022DETRsWC}     & 348M          & -      & -               &                               & 65.9\%     & 66.0\%      & -                & -                \\
InternImage-H \cite{wang2023internimage} & 2180M         & -      & -               &                               & 65.0\%     & 65.4\%      & -                & -                \\ \hline
RT-DETR-R50 \cite{rtdetr}                & 42M           & 136G   & 640             &      108(T4)                         & 53.1\%     &             &                  &                  \\
RT-DETR-R101 \cite{rtdetr}               & 76M           & 259G   & 640             &      74(T4)                         & 54.3\%     &             &                  &                  \\
RT-DETR-L \cite{rtdetr}                  & 32M           & 110G   & 640             &      114(T4)                         & 53.0\%     &             &                  &                  \\
RT-DETR-X \cite{rtdetr}                  & 67M           & 234G   & 640             &      74(T4)                         & 54.8\%     &             &                  &                 
\end{tabular}
\label{tab:sotacomparison}
\end{table*}


\section{Conclusions}

In this article, we introduced YotoR, a hybrid architecture that combines the Swin Transformer and YoloR models for object detection. YotoR outperforms its constituent models in both accuracy and inference speed. Although this study focused on four specific YotoR configurations, it opens avenues for exploring larger models and adaptations to new architectures beyond YoloR and Swin Transformer.

The fact that Transformer-based backbones can be combined with Yolo-like heads has broader implications, extending to the design of new hybrid models for different kinds of image-related tasks.

Further experiments are necessary to fully understand the impact of Yolo-style heads and implicit knowledge, as well as to address performance challenges in real-time processing.



Future research directions include upgrading YotoR's backbone, investigating implicit knowledge integration, exploring novel techniques for head enhancements, considering multi-modal networks, and incorporating language tasks into the predictions.
\clearpage  

%
%
\bibliographystyle{splncs04}

\end{document}